\documentclass[sigconf]{acmart}

\AtBeginDocument{%
  \providecommand\BibTeX{{%
    \normalfont B\kern-0.5em{\scshape i\kern-0.25em b}\kern-0.8em\TeX}}}

\usepackage{calligra}
\usepackage{algorithm}
\usepackage{algorithmicx}
\usepackage[noend]{algpseudocode} 
\usepackage{amsmath} 
\usepackage{graphicx}
\counterwithin{figure}{section}
\counterwithin{equation}{section}
\usepackage{subfigure}
\usepackage{url}
\usepackage{stfloats}
\usepackage{array}

\graphicspath{{figs/}}

\copyrightyear{2022} 
\acmYear{2022} 
\setcopyright{rightsretained} 
\acmConference[ICCAD '22]{IEEE/ACM International Conference on
Computer-Aided Design}{October 30-November 3, 2022}{San Diego, CA,
USA}
\acmBooktitle{IEEE/ACM International Conference on Computer-Aided
Design (ICCAD '22), October 30-November 3, 2022, San Diego, CA, USA}
\acmDOI{10.1145/3508352.3561112}
\acmISBN{978-1-4503-9217-4/22/10}

\begin{document}

\title{Factor Graph Accelerator for LiDAR-Inertial Odometry (Invited Paper)}

\author{Yuhui Hao$^*$}
\affiliation{%
  \institution{Tianjin University}
  \country{China}
}

\author{Bo Yu$^*$}
\affiliation{%
  \institution{PerceptIn}
  \country{U.S.A.}
}

\author{Qiang Liu$^\dagger$}
\affiliation{%
  \institution{Tianjin University}
  \country{China}
}

\author{Shaoshan Liu}
\affiliation{%
  \institution{PerceptIn}
 \country{U.S.A.}
\thanks{$^*$ indicates equal contribution to the paper.}
\thanks{$\dagger$ indicates the corresponding author of the paper.}
}

\author{Yuhao Zhu}
\affiliation{%
  \institution{University of Rochester}
  \country{U.S.A.}
}

\renewcommand{\shortauthors}{}


\newcommand{\website}[1]{{\tt #1}}
\newcommand{\program}[1]{{\tt #1}}
\newcommand{\benchmark}[1]{{\it #1}}
\newcommand{\fixme}[1]{{\textcolor{red}{\textit{#1}}}}
\newcommand{\answer}[1]{{\textcolor{blue}{\textit{#1}}}}

\newcommand*\circled[2]{\tikz[baseline=(char.base)]{
            \node[shape=circle,fill=black,inner sep=1pt] (char) {\textcolor{#1}{{\footnotesize #2}}};}}

\ifx\figurename\undefined \def\figurename{Figure}\fi
\renewcommand{\figurename}{Fig.}
\renewcommand{\paragraph}[1]{\textbf{#1} }
\newcommand{\figline}{{\vspace*{.05in}\hline}}

\newcommand{\Sect}[1]{Sec.~\ref{#1}}
\newcommand{\Fig}[1]{Fig.~\ref{#1}}
\newcommand{\Tbl}[1]{Tbl.~\ref{#1}}
\newcommand{\Equ}[1]{Equ.~\ref{#1}}
\newcommand{\Apx}[1]{Apdx.~\ref{#1}}
\newcommand{\Alg}[1]{Algo.~\ref{#1}}

\newcommand{\INPUT}{\item[\textbf{Input:}]}
\newcommand{\OUTPUT}{\item[\textbf{Output:}]}
\newcommand{\ALGORITHM}{\item[\textbf{Algorithm:}]}

\newcommand{\specialcell}[2][c]{\begin{tabular}[#1]{@{}c@{}}#2\end{tabular}}
\newcommand{\note}[1]{\textcolor{red}{#1}}

\newcommand{\proj}{\textsc{Archytas}\xspace}
\newcommand{\mode}[1]{\underline{\textsc{#1}}\xspace}
\newcommand{\sys}[1]{\underline{\textsc{#1}}}

\newcommand{\no}[1]{#1}
\renewcommand{\no}[1]{}
\newcommand{\RNum}[1]{\uppercase\expandafter{\romannumeral #1\relax}}

\def\cA{{\mathcal{A}}}
\def\cF{{\mathcal{F}}}
\def\cN{{\mathcal{N}}}

\def\bA{{\mathbf{A}}}
\def\bB{{\mathbf{B}}}
\def\bb{{\mathbf{b}}}
\def\bC{{\mathbf{C}}}
\def\be{{\mathbf{e}}}
\def\bH{{\mathbf{H}}}
\def\bJ{{\mathbf{J}}}
\def\bM{{\mathbf{M}}}
\def\bMi{{\mathbf{M}^{-1}}}
\def\bN{{\mathbf{N}}}
\def\bp{{\mathbf{p}}}
\def\bS{{\mathbf{S}}}
\def\bSp{{\mathbf{S'}}}
\def\bU{{\mathbf{U}}}
\def\bUi{{\mathbf{U}^{-1}}}
\def\bV{{\mathbf{V}}}
\def\bW{{\mathbf{W}}}
\def\bX{{\mathbf{X}}}
\def\bs{{\mathbf{s}}}
\def\bnd{{\mathbf{n_d}}}
\def\bnm{{\mathbf{n_m}}}


\begin{abstract}
Factor graph is a graph representing the factorization of a probability distribution function, and has been utilized in many autonomous machine computing tasks, such as localization, tracking, planning and control etc. We are developing an architecture with the goal of using factor graph as a common abstraction for most, if not, all autonomous machine computing tasks.
If successful, the architecture would provide a very simple interface of mapping autonomous machine functions to the underlying compute hardware.
As a first step of such an attempt, this paper presents our most recent work of developing a factor graph accelerator for LiDAR-Inertial Odometry (LIO), an essential task in many autonomous machines, such as autonomous vehicles and mobile robots.
By modeling LIO as a factor graph, the proposed accelerator not only supports multi-sensor fusion such as LiDAR, inertial measurement unit (IMU), GPS, etc., but solves the global optimization problem of robot navigation in batch or incremental modes.
Our evaluation demonstrates that the proposed design significantly improves the real-time performance and energy efficiency of autonomous machine navigation systems.
The initial success suggests the potential of generalizing the factor graph architecture as a common abstraction for autonomous machine computing, including tracking, planning, and control etc.
\end{abstract}





\maketitle

\section{Introduction}

Autonomous machine computing (AMC) is the next big trend in information technology in the coming decades, after personal computing, mobile computing, and cloud computing ~\cite{liu2021rise}. Specifically, AMC is the core technology stack that empowers various kinds of autonomous machines, including Mars or Lunar explorers, intelligent vehicles, autonomous drones, delivery robots, home service robots, agriculture robots, industry robots and many more that we have yet to imagine ~\cite{liu2021robotic}. 

Similar to other information technology stacks, the AMC technology stack consists of hardware, systems software and application software. Sitting in the middle of this technology stack is computer architecture, which defines the core abstraction between hardware and software. This abstraction layer allows software developers to focus on optimizing the software and hardware developers to focus on developing faster, more affordable, more energy-efficient hardware that can unlock the imagination of software developers. 

While there have been many recent proposals of computer architectures for AMC~\cite{liu2021daa,yu2020building,liu2017computer,liu2021archytas,gan2021eudoxus}, this paper is the first to explore factor graph as a common abstraction, or architecture, for autonomous machine operations. A factor graph is a graph representing the factorization of a probability distribution function and has been utilized in many autonomous machine operations, such as localization, tracking, planning and control~\cite{dellaert2017factor, dong2016motion}. As the initial step, we focus on developing a factor graph accelerator for LiDAR-inertial odometry, a key localization method in many autonomous machines. 

The rest of this paper is organized as follows. \Sect{sec:bac} introduces the background of factor graph. \Sect{sec:traits} summarizes the traits of factor graph computing that motivates the design of the proposed accelerator. \Sect{sec:hw} delves into the hardware design of the proposed accelerator. \Sect{sec:exp} presents the results of performance evaluation, and we conclude in \Sect{sec:conc}. 
\section{Background on Factor Graph}
\label{sec:bac}

Factor graphs are probabilistic graphical models that can essentially model complex state estimation problems. For representing a state estimation problem, the factor graph is organized as a bipartite graph consisting of variables connected by factors, where variables represent unknown states and factors connected to states represent probabilistic relations between states \cite{dellaert2021factor}.
According to the factor graph, the joint probability distribution of the entire states can be factorized into products of probabilistic functions; 
as a result, the state estimation is turned into the maximum a posterior (MAP) inference \cite{dellaert2017factor}, of which solution is to maximize the products of probabilistic factors.

Many AMC problems, such as estimation, planning and optimal control, have an optimization problem at their core, as a result factor graphs have been well applied to represent, reason and solve those AMC operations. The factor graph abstraction of robotic optimization brings several benefits: First, it is an unified model that well suits for various forms of optimization problem, such as filtering \cite{dellaert2012factor}, incremental smoothing \cite{indelman2012factor} and batch optimization \cite{chiu2013robust}. Second, it provides a concise and abstract programming interface to compose robotic optimization problems, through which all programmers need to program a robotic optimization application is to specify the variables, factor functions and their connections according to the graph. Third, its graph structure facilitates sparse data storage that can be exploited to optimize memory resources and performance. 

Several software libraries \cite{chiu2013robust,kummerle2011g,kaess2012isam2} using factor graph for robotic optimization have been developed. However, hardware accelerators for factor graph are less studied, which however are highly required by AMCs that usually have stringent power and performance constraints \cite{yu2020building}. To take the first step towards the hardware accelerator of factor graph, we use LiDAR-Inertial Odometry (LIO, a dominant localization approach for autonomous vehicles) \cite{shan2020lio} as an example to exploit characteristics of factor graph computing that facilitate hardware design. 
\section{Traits of Factor Graph Computing}
\label{sec:traits}


This section first introduces the general MAP inference formulation for factor graph 
(\Sect{sec:traits:map}). Then the algorithm for solving the LIO based on factor graph is illustrated in \Sect{sec:traits:sol}. The structural characteristic of the LIO factor graph that facilitates parallel computing is discussed in \Sect{sec:traits:para}. 

\subsection{MAP Inference for Factor Graph}
\label{sec:traits:map}


For AMC state estimation tasks, the core function is to compute unknown states $X$, such as poses (position and orientation) in localization applications, given the noisy measurements, $Z$. Factor graphs can essentially model AMC estimation problems, in which variable nodes represent unknown states and factor nodes represent functions that only apply on variables connected with the factor node. With the topology definition, a factor graph defines the factorization of a global function $\Phi(X)$. In AMC state estimation, the global function is a joint probability distribution. The object of state estimation is to maximize the joint probability, which turns into a MAP inference,


\begin{equation}
    X^{*}=\mathop{\arg\max}\limits_X\Phi(X)=\mathop{\arg\max}\limits_X\prod_i\phi_i(X_i)
    \label{equ:mapmax}
\end{equation}

Assuming that all factors are of the form as \Equ{equ:exp}, which include both Gaussian priors and likelihood factors derived from measurements corrupted by zero-mean, normally distributed noise, where $z_i$ is the $i$-th measurement or ground truth, $h_i(\cdot)$ is a measurement function that maps the state $X_i$ to be estimated to its sensor's measurement space, $\Sigma_i$ is the covariance matrix of the $i$-th measurement and $\Vert\cdot\Vert^2_{\Sigma_i}$ is the Mahalanobis norm that quantifies the error. Taking the negative log of \Equ{equ:mapmax} allows us to instead minimize a sum of nonlinear least-squares as \Equ{equ:mapmin}.

\begin{equation}
    \phi_i(X_i)\propto\exp{\Vert h_i(X_i)-z_i \Vert^2_{\Sigma_i}}
    \label{equ:exp}
\end{equation}
\begin{equation}
    X^{*}=\mathop{\arg\min}\limits_X\sum_i\Vert h_i(X_i)-z_i \Vert^2_{\Sigma_i}
    \label{equ:mapmin}
\end{equation}

Nonlinear least-squares problems can not be solved directly, but require an iterative solution starting from an initial estimate. Typical nonlinear solvers, such as the Gaussian-Newton method, have a similar compute process. They start from an initial $X^0$. In each iteration, an increment $\Delta$ is computed and applied to obtain the next estimate $X=X+\Delta$. This process stops when certain convergence criteria are reached, such as the change $\Delta$ falling below a small threshold. In each iteration, $\Delta$ is found by solving the linear least-squares problem as \Equ{equ:mapmatrix},
\begin{align}
    \Delta^*=\mathop{\arg\min}\limits_\Delta\sum_i\Vert A_{bi}\Delta_i-\epsilon_{bi}\Vert^2_2=\mathop{\arg\min}\limits_\Delta\Vert A_{b}\Delta-\epsilon_{b}\Vert^2_2
    \label{equ:mapmatrix}
\end{align}
\begin{equation}
    \begin{aligned}
    A_{bi}=\Sigma_i^{-\frac{1}{2}}\left.\frac{\partial h_i(X_i)}{\partial X_i}\right|_{X_i}, \quad
    \epsilon_{bi}=\Sigma_i^{-\frac{1}{2}}(z_i-h_i(X_i))
    \label{equ:whitening}
\end{aligned}
\end{equation}
where the Mahalanobis norm is transformed to 2-norm by \Equ{equ:whitening}. $A_b$ and $\epsilon_b$ are obtained by collecting all Jacobian matrices $A_{bi}$ and residuals $\epsilon_{bi}$ into a large matrix $A_b$ and right-hand-side (RHS) vector $\epsilon_b$, respectively.

In fact, the structure of factor graph is equivalent to the sparse pattern of $A_b$, i.e., each block row in $A_b$ corresponds to a factor node; each block column in $A_b$ corresponds to a variable node. Thus, the structure of factor graph directs the solution of \Equ{equ:mapmatrix}. By traversing all variable nodes in factor graph, the adjacency factors of each variable are combined and then factorized, namely variable elimination~\cite{dellaert2017factor}. After eliminating all variable nodes, \Equ{equ:mapmatrix} is transformed to \Equ{equ:qr}, where $R$ is an upper triangular matrix, also called the Bayes net in probabilistic graph theory~\cite{dellaert2021factor}, obtained by QR decomposition of $A_b$. Then $\Delta$ can be solved by back substitution in $R$.
\begin{align}
    \mathop{\arg\min}\limits_\Delta\Vert A_b\Delta-\epsilon_b \Vert^2_2&=\mathop{\arg\min}\limits_\Delta\Vert Q^TA_b\Delta-Q^T\epsilon_b \Vert^2_2 \nonumber\\
                                                                      &=\mathop{\arg\min}\limits_\Delta\Vert R\Delta-d \Vert^2_2
    \label{equ:qr}
\end{align}




However, data is obtained as a temporal sequence in many inference problems for AMC. The latency to solve grows over time to the point that real-time batch optimization is no longer feasible. In fact, the constraints formed by new measurements usually affect only a local part of factor graph and the remaining part is unchanged so that incremental smoothing requires solving only for partially affected factor graph~\cite{kaess2012isam2}. In terms of linear algebra, only partial entries in $R$ require to be updated~\cite{MichaelKaess2018}. Therefore, incremental smoothing can be regarded as the recalculation of a subgraph of the factor graph, of which process is consistent with the batch solution but the dimension is reduced significantly. 

\begin{figure}[t]
    \counterwithin{figure}{section}
    \centering
    \includegraphics[width=\columnwidth]{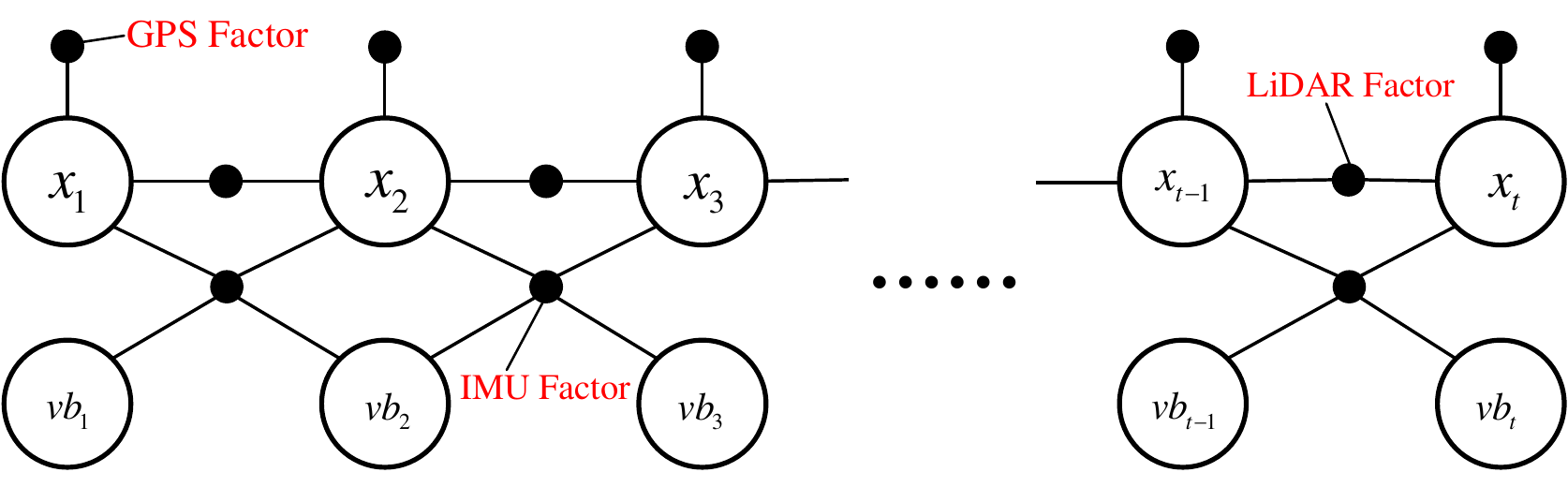}
    \caption{Factor graph of tightly coupled LiDAR-Inertial Odometry. The white circles represent the pose, velocity and biases variables. The black dots denote the different factors.} 
    \label{fig:liofg}
    \vspace{-0.2cm}
\end{figure}

\subsection{Solving for LIO Factor Graph}
\label{sec:traits:sol}

LIO uses GPS, IMU and LiDAR as the main sensors. GPS measurements generate constraints on each keyframe; IMU and LiDAR measurements produce constraints on two adjacent keyframes. Thus, the LIO factor graph has a chain-like structure, as shown in \Fig{fig:liofg}. The factor nodes contain the constraints formed by measurements of three sensors, and the variable nodes include the pose, velocity and biases of each keyframe. It results in a factor graph that is rich in regularity when solving for the MAP solution. The chain-like property avoids the necessity of traversing the factor graph, saving time and space overhead.

\begin{figure*}[htbp]
    \centering
    \subfigure[]{
        \label{fig:sve1}
        \includegraphics[width=0.6\columnwidth]{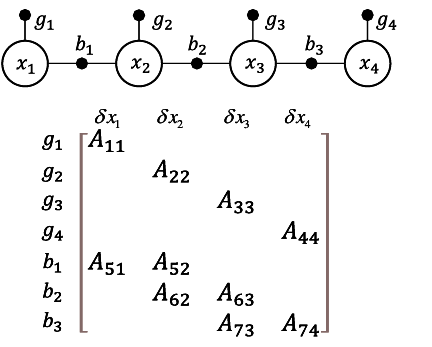}}
    \subfigure[]{
        \label{fig:sve2}
        \includegraphics[width=0.6\columnwidth]{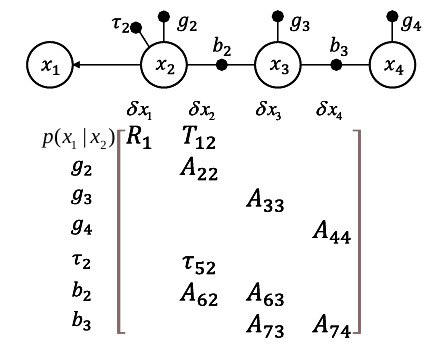}}
    \subfigure[]{
        \label{fig:sve3}
        \includegraphics[width=0.6\columnwidth]{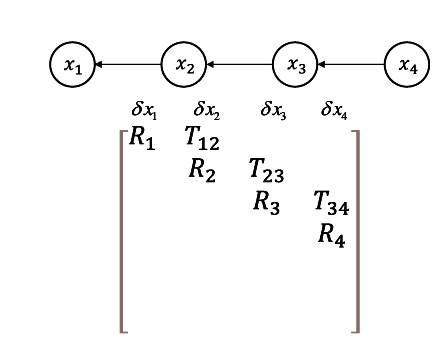}}
    \setlength{\abovecaptionskip}{0.cm}
    \caption{Factor graph and corresponding matrix obtained using serial elimination. (a) Toy example of LIO factor graph with four variables and its corresponding Jacobian matrix  $A_b$ in serial elimination order. (b) Factor graph and the corresponding matrix after eliminating $x_1$. (c) The Bayes net obtained using serial elimination and the corresponding upper triangular matrix $R$.}
    \label{fig:SVE}
\end{figure*}

\begin{algorithm}[t]
    \counterwithin{algorithm}{section}
    \caption{Variable Elimination On LIO Factor Graph}
    \label{alg:VELIO}
    \begin{algorithmic}[1]
        \Function{EliminationOnLio}{LIO Factor Graph $\Phi_{1:n}$}
        \For{$i=1$ to $n$}
         \State $\overline{A_i}$ $\gets$ $g_i$, $b_i$, $\tau_i$
        \State  $p(x_i|x_{i+1})$, $\tau_{i+1}$ $\gets$ partial QR decomposition on $\overline{A_i}$
        \EndFor
        \State\Return $p(x_1|x_2)p(x_2|x_3)\cdots p(x_n)$
        \EndFunction
    \end{algorithmic}
\end{algorithm}


\Fig{fig:sve1} shows the LIO factor graph for a toy example and its corresponding Jacobian matrix $A_b$, where the velocity and biases variables as well as the IMU factors are omitted for the sake of brevity and clarity. If the forward (serial) elimination order of $x_1\to x_2\to x_3\to x_4$ is selected, when eliminating $x_1$, the adjacent factors $g_1(x_1)$ and $b_1(x_1,x_2)$ are multiplied into the product $\psi(x_1,x_2)$ and factorized into a conditional probability $p(x_1|x_2)$ (the first block row in \Fig{fig:sve2}) and a new factor $\tau_2(x_2)$, that is
\begin{align}
    &\psi(x_1,x_2) \gets g_1(x_1)b_1(x_1,x_2) \label{prod} \\
    &p(x_1|x_2)\tau_2(x_2) \gets \psi(x_1,x_2)  \label{deco}
\end{align}
From the perspective of the matrix, \Equ{prod} corresponds to the process of constructing a matrix $\overline{A}$; \Equ{deco} represents the process of decomposing $\overline{A}$. When eliminating $x_1$, $\overline{A_1} = \begin{bmatrix} A_{11} & \\ A_{51} & A_{52} \end{bmatrix}$. Then perform partial $QR$ decomposition on $\overline{A_1} = Q_1\begin{bmatrix} R_{1} & T_{12}\\   & \tau_{52} \end{bmatrix}$, where $R_{1}$ is an upper triangular matrix, as shown in \Fig{fig:sve2}. The above process is then iterated for subsequent variables. As the number of iterations increases, the dimension of $\tau$ grows linearly, thus the dimension of $\overline{A}$ is also growing linearly. The obtained Bayes net is shown in \Fig{fig:sve3} after all variables are eliminated. When solving the MAP solution in back substitution, the reverse order of the elimination should be followed, i.e., $x_4\to x_3\to x_2\to x_1$ should be solved in turn. 

However, in each elimination, factors adjacent to $x_j$ to be eliminated are fixed due to the chain-like nature of LIO factor graph, i.e., only a GPS factor $g_j(x_j)$, a LiDAR factor $b_j(x_j,x_{j+1})$, and a new factor $\tau_j(x_j)$ obtained by previous elimination are involved when eliminating $x_j$. Therefore, the matrix $\overline{A_j}$ can be directly constructed without traversing the factor graph. Besides, during the back substitution, each variable depends only on its neighboring variable so that this process can also be performed without traversing the graph. \Alg{alg:VELIO} shows pseudocode for eliminating a LIO factor graph.

When solving LIO factor graph incrementally, a factor graph of the three most recent variables need to be recalculated according to~\cite{kaess2012isam2}, while the other parts remain unchanged. \Alg{alg:lio_incre} shows pseudocode for incremental inference on LIO factor graph.

\begin{algorithm}[t]
    \caption{Incremental Inference On LIO Factor Graph}    
    \label{alg:lio_incre}
    \begin{algorithmic}[1]
         \Function{IncrementalLio}{$g_{j+1}$, $b_{j}$}
        \State Reconstruct the factor graph $\Phi_{j-1:j+1}$ of $x_{j-1}, x_{j}$ and $x_{j+1}$
        \State E{\footnotesize LIMINATION}O{\footnotesize N}L{\footnotesize IO}($\Phi_{j-1:j+1}$)
        \State \Return $p(x_{j-1}|x_j)p(x_{j}|x_{j+1})p(x_{j+1})$
        \EndFunction
    \end{algorithmic}
\end{algorithm}
\vspace{-5pt}


\subsection{Parallel Computation of LIO Factor Graph}
\label{sec:traits:para}

The parallelism of elimination and back substitution is explored by considering the symmetric properties of LIO factor graph to improve the inference performance in hardware. Parallel elimination is feasible because only the elimination order is changed and the parts without data dependency are computed in parallel, while minimizing the filling-in (the number of non-zero entries) in the upper triangular matrix $R$ ~\cite{dellaert2021factor}. 

\begin{figure}[t]
    \centering
    \subfigure[]{
        \label{fig:pve2}
        \includegraphics[width=0.48\columnwidth]{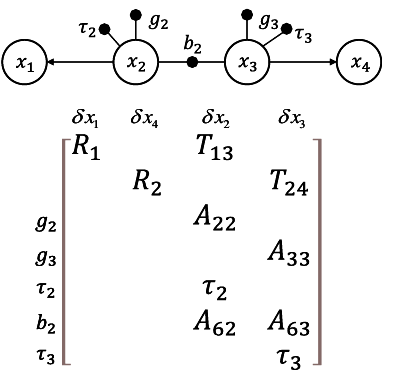}}
    \subfigure[]{
        \label{fig:pve3}
        \includegraphics[width=0.48\columnwidth]{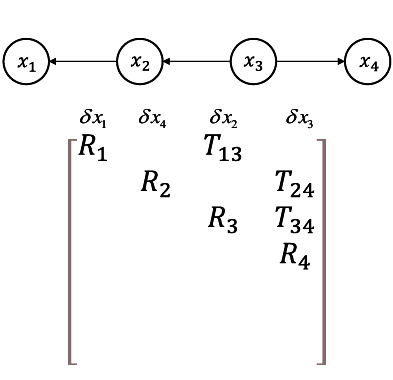}}
    \setlength{\abovecaptionskip}{0.cm}
    \caption{Factor graph and corresponding matrix obtained using parallel elimination. (a) Factor graph and the corresponding matrix after eliminating $x_1$ and $x_4$ simultaneously. (b) The Bayes net obtained using parallel elimination and the corresponding upper triangular matrix $R$. }
    \vspace{-10pt}
    \label{fig:PVE}
\end{figure}

Reviewing the toy example, when eliminating $x_1$, there is no common factor between $x_1$ and $x_4$. Therefore, the elimination of $x_1$ and $x_4$ is considered simultaneously. The computations involved are similar due to the symmetry of LIO factor graph, as shown in \Fig{fig:pve2}. Parallel elimination from both sides to the middle could continue if the chain was longer, but not in this case because $x_2$ and $x_3$ have common factor $b_2$ here. Moreover, the maximum dimension of the matrix $\overline{A_j}$ to be decomposed for parallel elimination is smaller than that for serial elimination, as the number of iterations required for parallel elimination is half that of serial elimination. Parallel elimination also results in a different Bayes net, in which case the root is in the middle and the directed edges point to the variables on both sides, as shown in \Fig{fig:pve3}. This means that the root variable can be solved first, after which the children variables on both sides can be solved in parallel in turn, e.g. $x_2$ and $x_4$ can be solved simultaneously.

The two elimination patterns get the same minimum filling-in, three in this toy example, with different elimination orders. However, the competitiveness of parallel elimination is reflected in: (a) parallelization of the matrix decomposition, (b) dimension reduction of the matrix to be decomposed and (c) parallelization of the back substitution.

Parallel elimination can also be employed for incremental smoothing which recalculates the symmetric LIO factor graph containing the three most recent variable nodes. Therefore, incremental smoothing of LIO factor graph is also accelerated by parallel elimination.

\section{Hardware Accelerator Design}
\label{sec:hw}

\begin{figure*}[t]
    \centering
    \includegraphics[width=2\columnwidth]{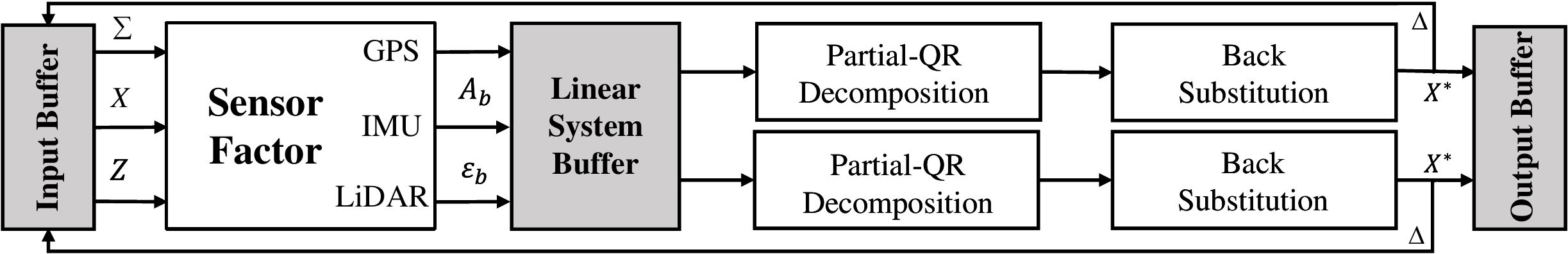}
    \caption{Overall hardware architecture.}
    \label{fig:hardware}
\end{figure*}


This section first gives an overview of the hardware architecture (\Sect{sec:hw:overview}). Then, how the key blocks in the hardware architecture are designed by leveraging the data and computation patterns inherent to LIO factor graph is presented, respectively (\Sect{sec:hw:reuse} to \Sect{sec:hw:qrd}).

\subsection{Hardware Design Overview}
\label{sec:hw:overview}
\Fig{fig:hardware} shows the hardware architecture of the accelerator, consisting of a series of optimized hardware blocks to accelerate the inference on LIO factor graph while minimizing the area and power consumption through circuit reuse. The main functional blocks of the accelerator include the factor block, the partial-QR decomposition blocks and the back-substitution blocks.

First, the state variables $X$, measurements $Z$ and covariance matrices $\Sigma$ are loaded from the input buffer, and then the residuals $\epsilon_{bi}$ and the Jacobians $A_{bi}$ are calculated by the factor block. The sensors supported by this hardware architecture are GPS, IMU and LiDAR. $A_b$ and $\epsilon_b$ used to construct the system of linear equations are stored in on-chip memory, which leverages the inherent sparsity in LIO factor graph to reduce memory size.

The hardware architecture uses two sets of partial-QR decomposition blocks and back-substitution blocks to accelerate the inference process.
After multiple iterations of partial-QR decomposition and back substitution, the increment $\Delta$ of the state variables is calculated. In each iteration, $\Delta$ is added to $X$ to obtain the updated state $X=X+\Delta$. If the convergence conditions are not met, $X$ is written directly to the input buffer for the next iteration. Otherwise, $X^*=X$ is output as the final result.
\vspace{-5pt}
\subsection{Circuit Reuse}
\label{sec:hw:reuse}


Computation similarities across algorithm modules are exploited to reuse circuits and thereby reduce the resource consumption. Circuit reuse here refers to two aspects: (1) Computation Results Reuse and (2) Functional Units Time-Multiplexing.



\begin{figure}[t]
    \begin{minipage}{0.48\columnwidth}
    \centering
    \includegraphics[width=\columnwidth]{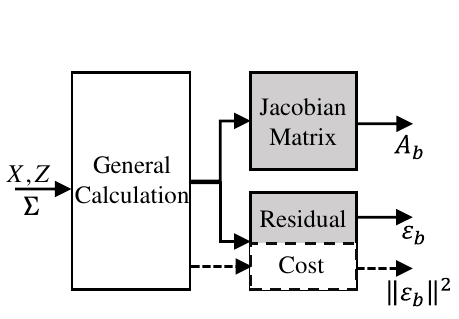}
    \caption{The data flow of the factor block. Solid lines and dashed lines denote the data flow to compute $\epsilon_b$, $A_b$ and the cost, respectively.}
    \label{fig:res_jac}
    \end{minipage}
    \hspace{3pt}
    \begin{minipage}{0.49\columnwidth}
    \centering
    \includegraphics[width=\columnwidth]{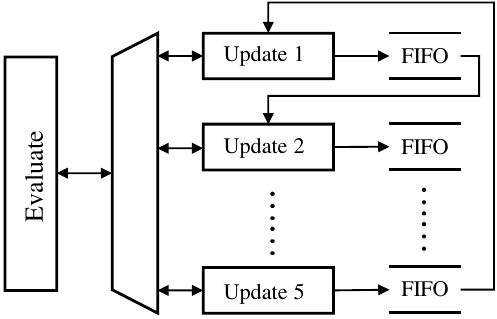}
    \caption{To ensure a balanced pipeline between Evaluate and Update, the QR decomposition block uses one Evaluate unit and multiple (five here) time-multiplexed Update units.}
    \label{fig:qr}
    \end{minipage}
    \\
     \vspace{-20pt}
\end{figure}

\paragraph{Computation Results Reuse} \Fig{fig:res_jac} shows the architecture of the factor block. There is a significant overlap of general parts in the three factors, which is divided into two levels: between sensors, and within sensors.  Although the IMU factor and LiDAR factor are produced by different sensors, they are both constraints on two adjacent keyframes, so there are a large number of same intermediate results in the computing process.  Computation overlap also exists in the IMU factor. While calculating the $\epsilon_{bi}$ in IMU, some items in $A_{bi}$ can be calculated at the same time. These common computations are finished by the General Calculation unit, avoiding repeated calculations.



\paragraph{Functional Units Time-Multiplexing} The factor block has two operating modes, which perform the calculation of $\epsilon_b$, $A_b$ and the cost function, respectively. 
$\epsilon_b$ and $A_b$ need to be constructed when solving $\Delta$ in each iteration. However, only $\epsilon_b^{new}$ is required to calculate the cost when evaluating whether accepting $\Delta$ or not. Therefore, the computations of $\epsilon_b$, $A_b$ and the cost are mapped onto the configurable basic computing units. In \Fig{fig:res_jac}, solid lines and dashed lines denote the data flow for calculating $\epsilon_b$, $A_b$ and the cost, respectively.

\vspace{-5pt}
\subsection{Storage Optimization}
\label{sec:hw:storage}

Storage optimization can be divided into three steps. The first step is to store $\epsilon_b$ and $A_b$ according to the sparse factor graph structure instead of the original matrix with substantial zero entries. Each factor needs to store its $\epsilon_{bi}$ and $A_{bi}$, in addition to its factor type (to know the  dimension of $\epsilon_{bi}$ and $A_{bi}$) and corresponding variable indexes. 


However, due to the chain structure of LIO factor graph, the sparsity pattern of the matrix $\overline A$ can be determined each time before variable elimination is performed. In other words, the factors and variables involved are fixed in each elimination so the factor types and corresponding variable indexes are always known. Therefore, in the second step, $\epsilon_{bi}$ and $A_{bi}$ corresponding to each factor can be stored sequentially without any redundant information. 

The third step is to skip the large number of zero and identity entries in $A_{bi}$, e.g. $A_{bi}$ in IMU factor has a large number of identity matrices and zero entries in fixed positions. Besides, since the partial derivatives of $\epsilon_{bi}$  with respect to $x_i$ and $x_{i+1}$ are symmetric in LiDAR factors, only half of them are stored. The Householder matrix constructed in the QR decomposition (\Sect{sec:hw:qrd}) can also take advantage of its symmetry to further reduce memory size.

\subsection{Optimization of QR Decomposition Block}
\label{sec:hw:qrd}

The partial-QR decomposition starts from the first column of the $m\times n$ matrix $\overline{A}$, from which the Householder matrix $H$ is constructed  (the Evaluate phase). $H\overline{A}$ zeros the entries below the diagonal of the first column, and updates the second to $n$-th columns (the Update phase). Then the obtained $(m-1)\times(n-1)$ matrix in the lower right corner is the input for the second iteration. The iteration continues until the $\hat{n}$-th column, where $\hat{n}$ represents the dimension of $x_i$ to be eliminated.

After each iteration, the entries below the diagonal are known to be zeros and they are not needed in subsequent calculations. Therefore, the calculations of these zeros entries can be omitted directly so as to improve performance and save resources.

\Fig{fig:qr} shows the architecture of the block, where each Evaluate-Update pair performs one iteration of the decomposition. The analysis of the fine-grained data dependencies shows that the Evaluate-Update phase can be pipelined. The Evaluate phase of next iteration and the Update phase of current iteration can start at the same time. However, the Update phase is usually more time-consuming than the Evaluate phase. Therefore, this block designs an Evaluate unit and $n_u$ time-multiplexed Update units. The larger the $n_u$, the better the performance, but the larger the area. 


\vspace{-0.1cm}
\section{Performance Evaluation}
\label{sec:exp}

\begin{figure*}[t]
    \begin{minipage}[t]{0.8\columnwidth}
        \centering
        \setlength{\abovecaptionskip}{0.cm}
        \includegraphics[width=\columnwidth]{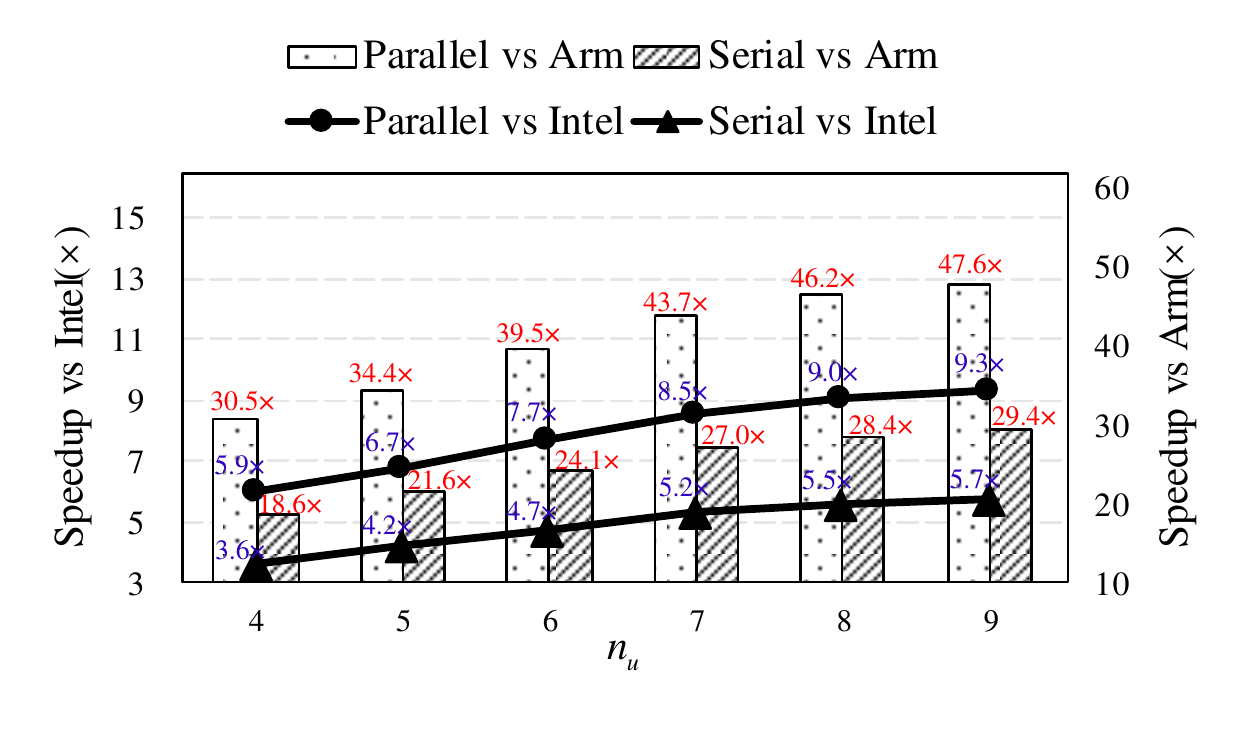}
        \caption{With the increase of $n_u$, the speedup of the accelerator working in parallel elimination and serial elimination modes compared with Intel and Arm.}
        \label{fig:speedup}
    \end{minipage}
    \hspace{3pt}
    \begin{minipage}[t]{0.64\columnwidth}
        \centering
        \setlength{\abovecaptionskip}{0.cm}
        \includegraphics[width=\columnwidth]{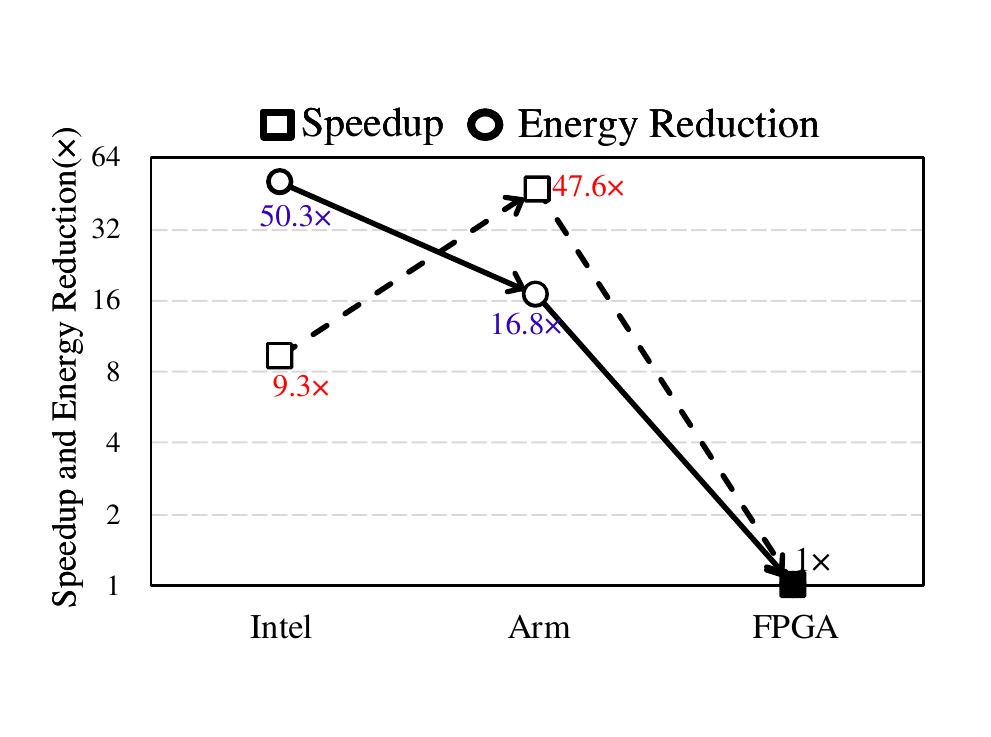}
        \caption{Energy reduction compared to Intel and Arm when the accelerator works at its best performance.}
        \label{fig:energy}
    \end{minipage}
    \hspace{3pt}
    \begin{minipage}[t]{0.55\columnwidth}
        \centering
        \setlength{\abovecaptionskip}{0.cm}
        \includegraphics[width=1\columnwidth]{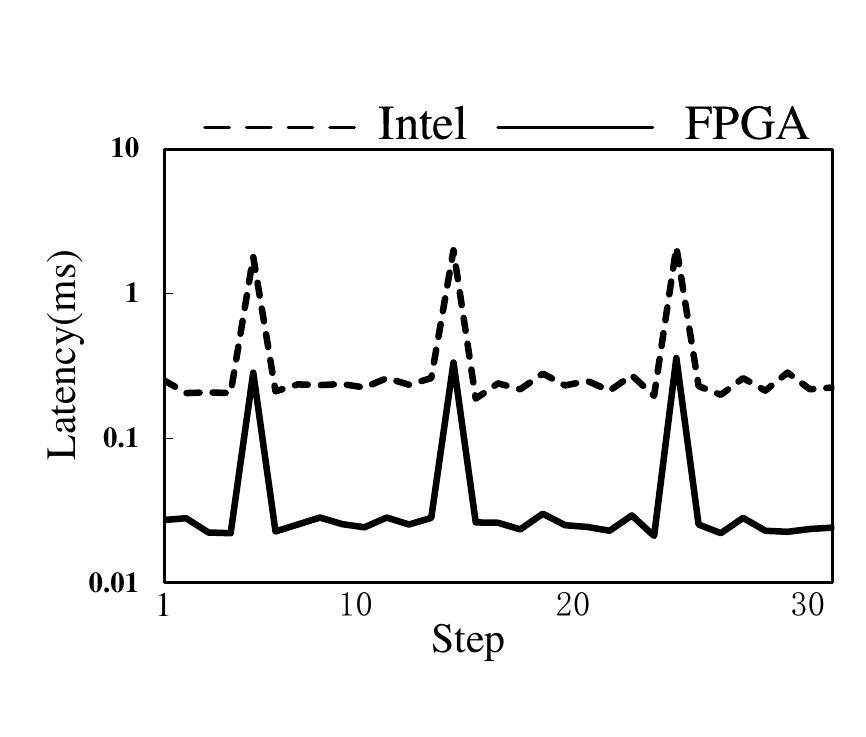}
        \caption{Latency comparison between the accelerator and Intel for factor graph optimization when running the Park dataset.}
        \label{fig:run}
    \end{minipage}
    \vspace{-15pt}
\end{figure*}

To evaluate the proposed accelerator, we conducted a series of experiments. We first introduce the experimental setup (\Sect{sec:exp:setup}). Then the results are presented in three parts. The first part shows the speed and energy of the proposed hardware accelerator compared to software (\Sect{sec:exp:perf}). The second part gives the localization accuracy of the accelerator while running the datasets (\Sect{sec:exp:acc}). The third part performs some comparisons before and after we optimize the hardware based on the data and computation patterns inherent to LIO factor graph (\Sect{sec:exp:res}).

\subsection{Experimental Setup}
\label{sec:exp:setup}
The accelerator is designed using Vitis-HLS, and synthesized and implemented onto the Xilinx Zynq-7000 SoC ZC706 FPGA using Vivado Design Suite 2021.1. The accelerator operates at a fixed frequency of 143 MHz. The FPGA power consumption is estimated by the Vivado power analysis tool using real workloads under test. All power and resource consumption data are obtained after the design passes the post-layout timing. The accelerator is evaluated on two common datasets: the Walking \cite{walking} and the Park \cite{park}. The Walking dataset was collected using a custom-built handheld device on the MIT campus. The Park dataset was collected in a park covered by vegetation, using an unmanned ground vehicle.

A software implementation of localization is used as a baseline, which uses GTSAM \cite{gtsam} to implement factor graph optimization for sensor fusion in Robotics. The software is evaluated on two hardware platforms: the one on the 11th Intel processor that has 16 cores and operates at 2.5 GHz, and the other on the quad-core Arm Cortex-A57 processor on the Nvidia mobile Jetson TX1 platform \cite{tx1} operating at 1.9 GHz. The Intel CPU power is measured through a power meter and the Arm core power is measured through the power sensing circuitry on TX1.

\vspace{-0.1cm}
\subsection{Performance Evaluation}
\label{sec:exp:perf}
We first evaluate the performance and energy efficiency of the accelerator. The proposed accelerator with FLP32 is compared to the Intel CPU implementation and the Arm implementation with FLP64.

By changing the number of Evaluate-Update units in the QR decomposition blocks, we obtain multiple sets of circuits with different performance. \Fig{fig:speedup} shows the speedup of the accelerator with different configurations including $n_u$ and elimination mode over Intel and Arm. The results show that the best performance design achieves 9.3$\times$ speedup over Intel and 47.6$\times$ speedup over Arm. Performing parallel elimination can improve performance by an average of 1.6$\times$ compared to serial elimination.  \Fig{fig:energy} demonstrates the energy efficiency. It shows that the design with the highest performance has an energy reduction of 50.3$\times$ over Intel and 16.8$\times$ over Arm. \Fig{fig:run} shows the latency of the accelerator compared to Intel while running the Park dataset. Batch optimization of LIO factor graph is performed at steps 5, 15, and 25, and incremental smoothing is performed for the rest.

\vspace{-0.1cm}
\subsection{Localization Accuracy Evaluation}
\label{sec:exp:acc}
\begin{figure*}[ht]
    \begin{minipage}[t]{0.7\columnwidth}
        \centering
        \setlength{\abovecaptionskip}{0.cm}
        \includegraphics[width=\columnwidth]{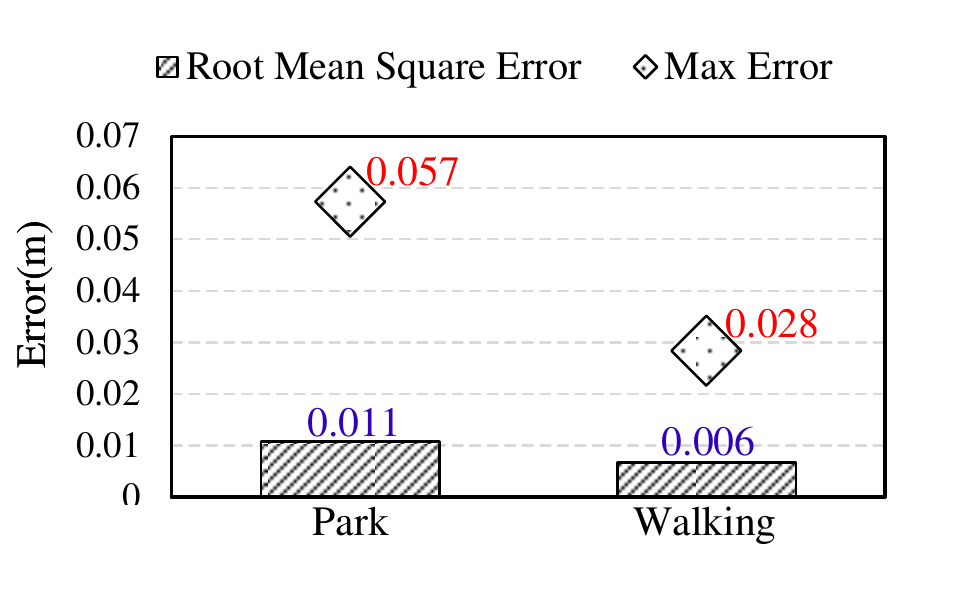}
        \caption{localization accuracy measured by Maximum Error and Root Mean Square Error.}
        \label{fig:accuracy}
    \end{minipage}
    \hspace{3pt}
    \begin{minipage}[t]{0.71\columnwidth}
        \centering
        \setlength{\abovecaptionskip}{0.cm}
        \includegraphics[width=1\columnwidth]{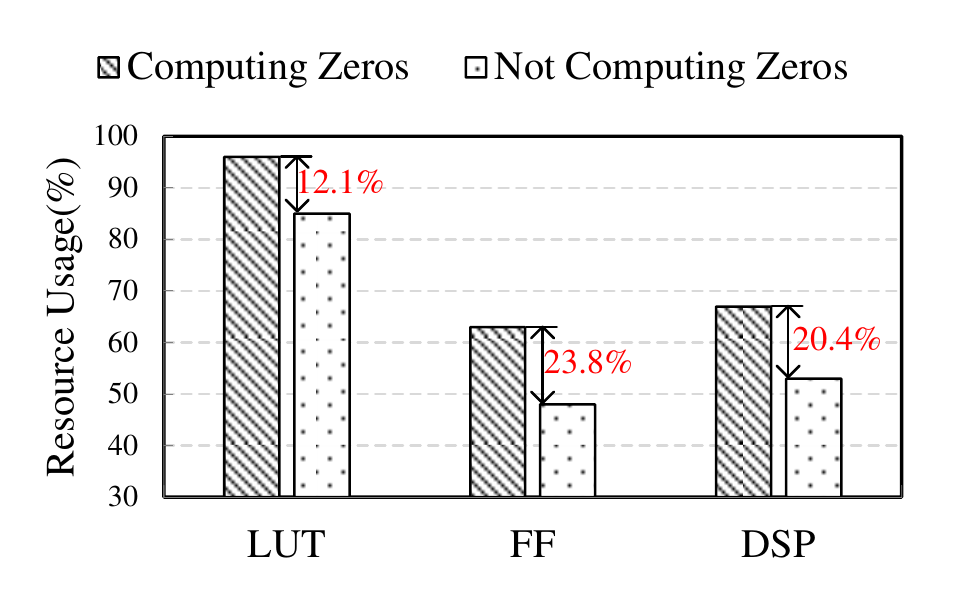}
        \caption{Resource savings from omitting the computation of zeros.}
        \label{fig:zeros}    
    \end{minipage}
    \hspace{3pt}
    \begin{minipage}[t]{0.58\columnwidth}
        \centering
        \setlength{\abovecaptionskip}{0.cm}
        \includegraphics[width=1\columnwidth]{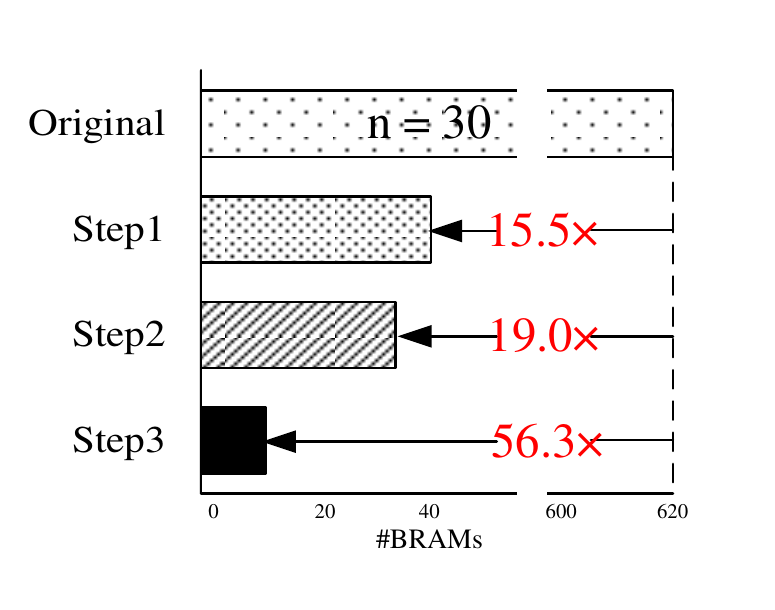}
        \caption{Matrix $A_b$ Storage optimization.}
        \label{fig:bram}    
    \end{minipage}
    \vspace{-10pt}
\end{figure*}

With the improvement in performance and energy reduction, we are also concerned about the localization accuracy of the accelerator, as illustrated in \Fig{fig:accuracy}. The localization accuracy is evaluated in terms of two Relative Pose Errors (RPEs): Root Mean Square Error (RMSE) and Maximum Error (ME). The result shows that our accelerator achieves the high localization accuracy compared to the ground truth. On the Walking dataset, its RMSE and ME are 0.6cm and 2.8cm, respectively. On the Park dataset, its RMSE and ME are 1.1cm and 5.7cm, respectively.

\renewcommand\arraystretch{0.7}
\begin{table}[t]
    \centering
    \counterwithin{table}{section}
    \caption{FPGA resource consumption (utilization percentages and absolute numbers) for parallel and serial elimination modes.}
    \label{tbl:resource}
    \begin{tabular}{c|cccc}
        \toprule[0.15em]
        \textbf{Mode} & \textbf{LUT}      & \textbf{FF} & \textbf{BRAM} & \textbf{DSP} \\
        \midrule[0.05em]
        Parallel      & \specialcell{85\%\\(184990)} & \specialcell{48\%\\(211895)} & \specialcell{15\%\\(163)} & \specialcell{53\%\\(473)} \\
        Serial        & \specialcell{46\%\\(99733)} & \specialcell{28\%\\(120416)} & \specialcell{8\%\\(88)} & \specialcell{28\%\\(255)}  \\
        \bottomrule[0.15em]
    \end{tabular}
    \vspace{-0.4cm}
\end{table}

\subsection{Resource Usage Evaluation}
\label{sec:exp:res}
The resource consumption of the best performance design with two elimination modes is shown in \Tbl{tbl:resource}. The results show that our design is memory-friendly, but consumes a lot of FFs and LUTs. This is because we store part of the intermediate data on FFs and LUTs instead of BRAMs to improve the performance.

The resources that can be saved by omitting the zeros computation in partial-QR decomposition blocks are shown in \Fig{fig:zeros}, and the results indicate that it saves 12.1\% LUTs, 23.8\% FFs, and 20.4\% DSPs in the best performance design.

In \Sect{sec:hw:storage} we optimize the storage according to the chain structure of LIO factor graph. \Fig{fig:bram} presents the results of memory saving. It shows that the memory size drops by 15.5$\times$, 19.0$\times$ and 56.3$\times$ after three steps of optimization with 30 keyframes, respectively.
\vspace{-0.5cm}
\section{Conclusion}
\label{sec:conc}

Rise of the autonomous machines demands an effective and efficient computer architecture to provide a concise and precise abstraction of the underlying autonomous machine operations, so as to unlock the imagination of AMC application developers. We believe a great candidate for AMC computer architecture is factor graph, which is a graph representing the factorization of a probability distribution function, and has been utilized in many autonomous machine computing functions, such as localization, tracking, planning and control etc. 

This paper presents the first work exploring factor graph architecture for autonomous machine computing, starting with LiDAR-Inertial Odometry, a key method in autonomous machine localization. Through exploiting the traits of factor graph computing, we have achieved up to 9$\times$ acceleration of autonomous machine localization compared to an advanced Intel CPU, along with 50$\times$ improvement of energy efficiency. Based on this initial success, we are going to generalize the factor graph architecture to provide computing support for other autonomous machine functions, including tracking, planning, and control etc.
\vspace{-5pt}

\begin{acks}
The authors would like to thank the support of the National Natural Science Foundation of China under Grant U21B2031.
\end{acks}

\bibliographystyle{unsrt}
\bibliography{references.bib}

\end{document}